\documentclass[conference]{IEEEtran}
\IEEEoverridecommandlockouts
\usepackage{cite}
\usepackage{amsmath,amssymb,amsfonts}
\usepackage{algorithmic}
\usepackage{graphicx}
\usepackage{textcomp}
\usepackage{subfiles}
\usepackage{xcolor}
\usepackage{hyperref} 

\def\BibTeX{{\rm B\kern-.05em{\sc i\kern-.025em b}\kern-.08em
    T\kern-.1667em\lower.7ex\hbox{E}\kern-.125emX}}
\begin{document}

\title{LLM-based Interactive Imitation Learning for Robotic Manipulation}

\author{
    Jonas Werner, Kun Chu$^{*}$, Cornelius Weber, Stefan Wermter\\
    Knowledge Technology, Department of Informatics\\
    University of Hamburg, Hamburg, Germany\\
    \texttt{Tubicor@outlook.de}, \texttt{\{kun.chu,cornelius.weber,stefan.wermter\}@uni-hamburg.de}
    \thanks{$^{*}$Corresponding author.}
}
\maketitle

\begin{abstract}
Recent advancements in machine learning provide methods to train autonomous agents capable of handling the increasing complexity of sequential decision-making in robotics. Imitation Learning (IL) is a prominent approach, where agents learn to control robots based on human demonstrations. However, IL commonly suffers from violating the independent and identically distributed (i.i.d) assumption in robotic tasks. Interactive Imitation Learning (IIL) achieves improved performance by allowing agents to learn from interactive feedback from human teachers. Despite these improvements, both approaches come with significant costs due to the necessity of human involvement. Leveraging the emergent capabilities of Large Language Models (LLMs) in reasoning and generating human-like responses, we introduce LLM-iTeach --- a novel IIL framework that utilizes an LLM as an interactive teacher to enhance agent performance while alleviating the dependence on human resources. Firstly, LLM-iTeach uses a hierarchical prompting strategy that guides the LLM in generating a policy in Python code. Then, with a designed similarity-based feedback mechanism, LLM-iTeach provides corrective and evaluative feedback interactively during the agent's training. We evaluate LLM-iTeach against baseline methods such as Behavior Cloning (BC), an IL method, and CEILing, a state-of-the-art IIL method using a human teacher, on various robotic manipulation tasks. Our results demonstrate that LLM-iTeach surpasses BC in the success rate and achieves or even outscores that of CEILing, highlighting the potential of LLMs as cost-effective, human-like teachers in interactive learning environments. We further demonstrate the method's potential for generalization by evaluating it on additional tasks. The code and prompts are provided at: \href{https://github.com/Tubicor/LLM-iTeach}{https://github.com/Tubicor/LLM-iTeach}.
\end{abstract}


\section{Introduction}

In a time when robotic assistance is growing in popularity, the ability to program such systems in sequential decision-making tasks holds great potential. As robotic environments get increasingly more complex, hard-coding specific behaviors for robotic manipulation is becoming less feasible, making it essential to develop autonomous agents that can learn and adapt. Machine Learning (ML) offers dynamic algorithms that enable an agent to learn to derive a policy from data\cite{murphy2012machine, RLIntro}. One prominent approach in ML is Imitation Learning (IL) \cite{AlgoImitationLearning} which offers a fast way to train intelligent agents for complex tasks \cite{AlgoImitationLearning}. Prominent IL methods like Inverse Reinforcement Learning \cite{russell1998learning} or Behavior Cloning \cite{bain1995framework} use a set of human-provided demonstrations to induce the desired behavior into the agent.

\begin{figure}
    \vspace*{4mm}
    \centering
    \includegraphics[width=0.95\linewidth]{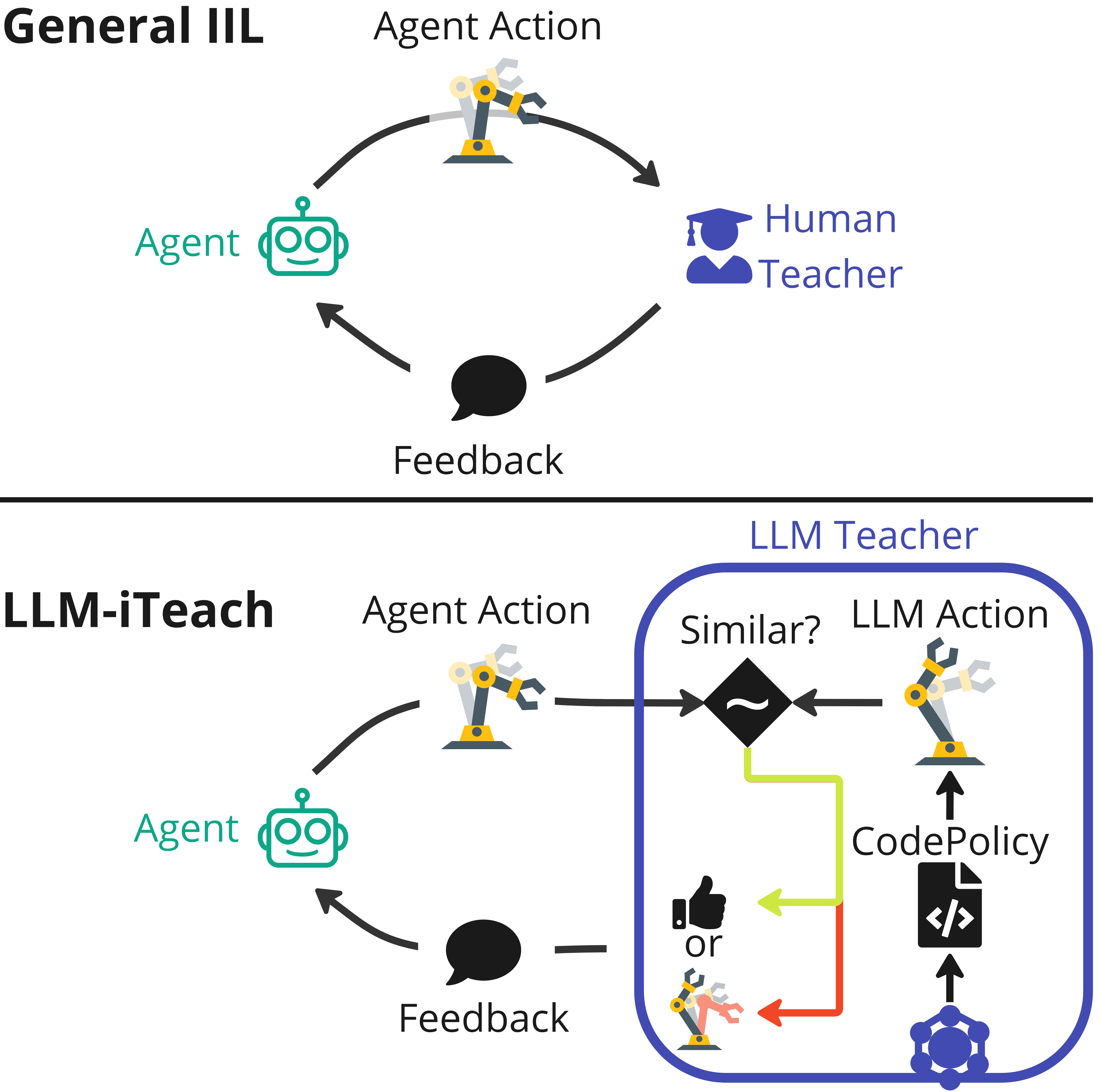}
    
    \caption{Comparison of the General IIL framework and our proposed LLM-iTeach. In IIL, a human teacher observes the agent's actions and provides timely feedback. In contrast, LLM-iTeach first prompts the LLM to encapsulate its reasoning hierarchically into CodePolicy, and then provides feedback to the agent in an evaluative or corrective manner through the similarity-checking mechanism designed in LLM-iTeach.}
    
    \label{fig:llm-teach}

\end{figure}
However, IL suffers from compounding errors and data-mismatch issues due to violating the common i.i.d. assumption in sequential decision-making problems \cite{DAGGER}. Interactive Imitation Learning (IIL) enhances IL by mitigating these negative impacts by enabling the agent to learn from human feedback on its actions during its interactions with the environment \cite{interactiveimitationlearningrobotics}. 
The teacher imparts knowledge on how to solve the task in its feedback, and the agent adapts its behavior accordingly, gradually acquiring the skill from the teacher. 
The IIL framework CEILing \cite{CEILing} proposes to integrate versatile feedback from a human teacher, including evaluative and corrective feedback, into the IL agent during the training, yielding state-of-the-art performance in robotic manipulation tasks. 
While CEILing renders good performance, it relies on human supervision throughout the training phase, which is usually time-consuming and costly. 
To avoid this, a way to automatically supervise the agent's learning and provide timely feedback would be required.

With the recent advancements in prompting techniques and increasing scales, LLMs offer reasoning capabilities and demonstrate common knowledge in a way that mimics human understanding \cite{wei2023chainofthoughtpromptingelicitsreasoning, LLMZeroShot}. LLMs have been applied in the 3D environment of robotic tasks for planning \cite{VoxPoser, ExtractActionForAgents, chu2024large, chu2025llmmap, Zhao2023}, generating \cite{LLMTrajectoryGeneratorWithCode, ExploRLLM} and judging actions or trajectories \cite{CodeAsReward, chu2023accelerating}. Recently, LLMs have proven effective in IIL as support for the human teacher in converting human verbal corrections to robot-understandable actions \cite{OLAF}. Given the capabilities of LLMs, their promising adaptation as support in IIL, and their potential to serve as a more cost-effective alternative to human teachers, the research question arises: \emph{Can LLMs interactively teach an imitation learning agent in robotic manipulation by providing corrective and evaluative feedback?}

To address the research question, this work proposes \textbf{LLM-iTeach} (\textbf{LLM i}nteractive \textbf{T}eacher giving \textbf{e}valuative \textbf{a}nd \textbf{c}orrective \textbf{h}uman-like feedback), which utilizes an LLM as its teacher to train an agent in robotic manipulation. The LLM is prompted to generate a policy for the task's sequential decision-making process, which we term CodePolicy. The CodePolicy is generated using a hierarchical prompting mechanism adapted from \cite{CaP} and indirectly provides the LLM's interactive feedback to teach the agent. To evaluate the performance of LLM-iTeach and the LLM's teaching capabilities\, we compare LLM-iTeach to the baselines Behavior Cloning (BC) \cite{AlgoImitationLearning}, a classical IL method, and CEILing \cite{CEILing}, a state-of-the-art IIL method employing a human teacher. To attain comparability, we keep the same experimental setup with CEILing and provide LLM-generated demonstrations for BC during the experiments. In addition, we extended the LLM-iTeach to four typically complex tasks, showcasing its ability to extend it to new tasks, compared with learning from human teachers' feedback. The experimental results demonstrate the effectiveness of LLM-iTeach and present the LLM as a viable alternative to a human teacher. 

The main contributions of this work are threefold. Firstly, we proposed the LLM-iTeach framework that interactively teaches an IL agent to perform complex robotic manipulation tasks, achieving comparable performance to the state-of-the-art IIL method. Secondly, the adapted hierarchical prompting strategy allows an LLM to generate a task-specific policy, and based on that, evaluative or corrective feedback according to a similarity-checking mechanism is continuously generated during the agent's training. Thirdly, we extend the LLM-iTeach framework to more complex tasks, showcasing that it is adaptive to new task environments with a simple description of the tasks.  


\section{Preliminaries}
\paragraph{Markov Decision Process}
An MDP \cite{MDP} is a framework that can formalize the sequential decision-making problem by modeling it as a tuple $\mathcal{<S, A, T, R, \gamma>}$ with four components: The set of all possible states is denoted by $ \mathcal{S}$, while $\mathcal{A}$ represents the set of all possible actions. The transition function, $\mathcal{T}: \mathcal{S} \times \mathcal{A} \rightarrow \mathcal{S}$, defines the transition to a state $s' \in \mathcal{S}$ given that action $a \in \mathcal{A}$ is taken in state $s\in \mathcal{S}$. Additionally, the reward function, $\mathcal{R}: \mathcal{S} \times \mathcal{A} \rightarrow \mathbb{R}$, determines the reward $r \in \mathbb{R}$ provided when action $a$ is executed in state $s$. The discount factor $\gamma \in [0, 1]$ determines the importance of future rewards relative to immediate rewards.

The agent interacts with the environment and, at each time step $t$, it receives the observation of the environment in terms of state $\mathit{s_t}$. Then it determines an action following a policy $\pi$ as below,

\begin{equation}\label{eq:Agent}
    Agent := \pi: \mathcal{S}\rightarrow \mathcal{A}.
\end{equation}


The duration between the start and end of the task is called an episode. The sequence of states and actions encountered by the agent during an episode is known as a trajectory, denoted as $\tau = (s_0,a_0,...,s_T,a_T)$, where \(\mathit{T}\) represents the total number of time steps the agent undergoes. The episode ends either upon task completion or when a predefined stop condition, such as a maximum number of steps, is met. In sequential decision-making, the goal is to find the policy $\mathit{\pi^*}$ that maximizes an objective function $J(\pi)$ often related to the reward $\mathcal{R}$. 




\paragraph{Imitation Learning} IL is an approach to adapting an agent to a sequential decision-making task in a robotic setting in which a human demonstrates accomplishing the task \cite{AlgoImitationLearning, interactiveimitationlearningrobotics}.
In general, the way that the teacher executes an action $a$ by receiving the state of the environment $s$ can be denoted as the teacher function $H$ as follows,

\begin{equation}\label{eq:ILTeacher}
    Teacher := H: \mathcal{S}\rightarrow \mathcal{A}.
\end{equation}

The objective function $J(\pi)$ is not directly accessible in IL, because it is implicitly given in the teacher's reasoning. However, the optimal policy $\pi^*$, which reflects the teacher's reasoning, can be determined by its similarity to the teacher's function $H$. As a result, the objective of the decision-making problem is to find the policy that maximizes the similarity. This is achieved by minimizing the loss function $L$, which measures the difference between the policy and the teacher's behavior.

\begin{equation}\label{eq:optimIL}
    \pi^* = \arg\min_{\pi}L(\pi,H)
\end{equation}

A common IL method is Behavior Cloning (BC) \cite{AlgoImitationLearning}, which approximates the policy $\pi*$ from the limited number of demonstrations provided by human teachers. 



However, IL faces the problem of compound errors and distribution shifts due to violating the i.i.d assumption in sequential decision-making, leading to unsatisfactory performance \cite{DAGGER, ross2010efficient}. Because the policy is learned on a set of demonstrations to predict the teacher's behavior and these predictions affect future states, it violates the independence of training and test data generally assumed in machine learning. 



\section{Related Works}

\paragraph{Interactive Imitation Learning} \label{ch:IIL} IIL is a class of methods that learns from the knowledge provided by a teacher in a sequential loop. The teacher provides feedback intermittently during the agent's execution of the task. The difference to IL is that in IIL the actions executed on the robot during the training are generated by following the agent's policy. This reduces compounding errors and distribution shift issues since more complete data is incrementally collected \cite{DAGGER, interactiveimitationlearningrobotics}. While the object remains the same, a key distinction from IL is that the teacher function maps the sate-action pairs to a certain form of human feedback, which is not limited to actions in IIL. The output of $\mathcal{H}$ is termed feedback to emphasize that, in the IIL approach, the agent controls the robot, not the teacher. The feedback can be of various types and may be provided sporadically, which can be classified into corrective or evaluative \cite{interactiveimitationlearningrobotics, CMNW18, chernova2022robot}. IIL methods in general let teachers train agents that perform better than policies obtained with standard IL \cite{interactiveimitationlearningrobotics}. One example of IIL is CEILing\cite{CEILing}, which provides for an interaction between the agent controlling the robot and the human, while simultaneously training the agent's policy with the given feedback. To be more specific, CEILing provides a pipeline to integrate human corrective and evaluative feedback in IL, obtaining state-of-the-art performance in robotic manipulation tasks. There has not yet been an IIL method without a human as a teacher. This work introduces an IIL framework that replaces the human teacher by an LLM teacher.

\paragraph{LLMs in Robotics} LLMs hold great potential to minimize the need for human resources with the widespread recognition as common-sense reasoners \cite{AttentionAllNeed, raffel2023exploring, SystemSurveyPromptMethods} in providing high-level instructions in robotics. Several works have shown that LLMs can process low-level data like positional vectors, object distances, or trajectories in robotic environments \cite{OLAF, chu2023accelerating, LLMTrajectoryGeneratorWithCode, ProgPrompt}. Yet generating low-level actions is harder for LLMs than abstracted representations \cite{LanguagetoReward, ExploRLLM, CaP}. LLMs excel in robotic tasks by generating code, allowing an agent to reason and plan like humans, to create actions \cite{ExtractActionForAgents, GPTPromptDesign, CaP, LanguagetoReward, ProgPrompt}. In the context of code generation, APIs provide further abstraction by offering complex functions unknown to the LLM \cite{CaP, GPTPromptDesign, LLMTrajectoryGeneratorWithCode, ExtractActionForAgents}. A key technique in prompting an LLM is chain-of-thought (CoT), which instructs it to break down answers into steps, enhancing performance for both zero-shot and few-shot learning \cite{LLMTrajectoryGeneratorWithCode, LLMZeroShot, CaP, MathPrompter, zhao2024enhancing}. 
Hierarchical prompting introduced in \cite{CaP} is a semi-automatic design principle to enable the LLM to generate to break down complex tasks into smaller steps through layered prompts, where each prompt can generate specific functions that invoke other prompts recursively.
The hierarchical structure allows for individual reaction to perceptual inputs and is applied for its flexibility and robustness in problem formulation in other works like VoxPoser \cite{VoxPoser}. Hierarchical prompting has not been applied yet in the construction of corrective feedback in IIL. With this work we propose a framework that leverage the abilities of hierarchical prompting to create such feedback.


\section{Approach}\label{approach}
To address the research question of to what degree an LLM can interactively teach an IL agent in robotic manipulation, this work proposes LLM-iTeach, which enables the LLM to provide corrective and instructive feedback to bootstrap the imitation learning while freeing up human effort. 

\subsection{LLM Teacher}
\paragraph{Hierarchical Prompting} 
LLMs can provide actions and plans for robotic tasks, yet they are faced with their high inference time when directly rendering those, which poses a problem for frequent interaction in IIL. To circumvent this, we designed LLM-iTeach to encapsulate the LLM's reasoning into a generated code termed CodePolicy. This generation is done before the interactions of the learning phase. Additionally, an executable policy providing an action is also less prone to noise than prompting an LLM directly. 

\begin{figure}[!t]
    \centering
    \includegraphics[width=0.6\linewidth]{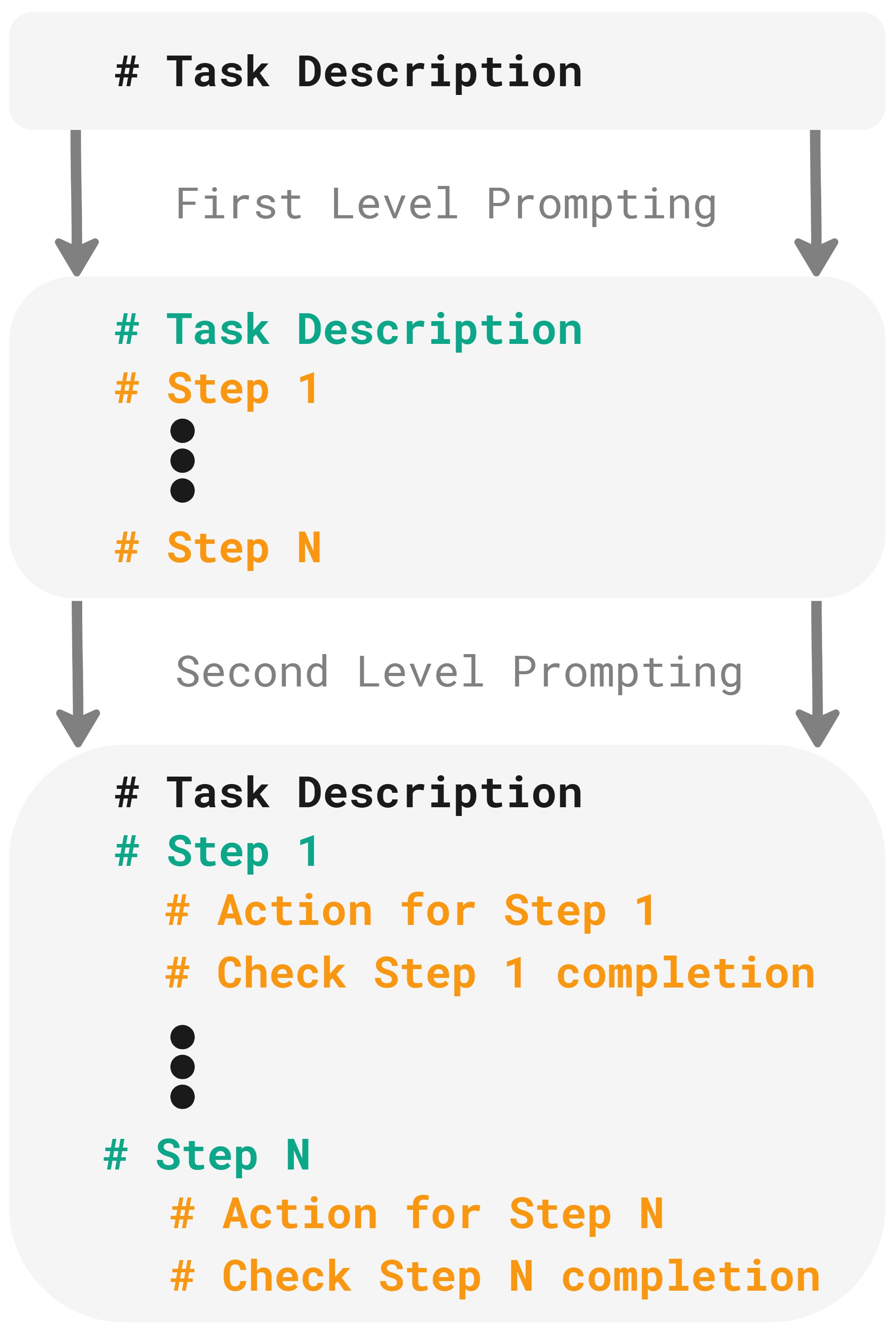}
    \caption{Abstraction of consecutive construction of CodePolicy with hierarchical prompting. The green colored lines are the instructions of the prompts, and the orange lines are the corresponding generation of the LLM. The two hierarchical levels of the prompting build on top of each other and are composed of the CodePolicy.}
    \label{fig:CodePolicy}
\end{figure}

During learning, the LLM's preferred action given the state of the environment is indirectly determined by the execution of the CodePolicy. The Generation of the CodePolicy is done before the learning of the agent, by prompting the LLM hierarchically (as shown in Fig. \ref{fig:CodePolicy}) on two levels: 

\begin{itemize}
    \item At the first level, the LLM receives a description of the task consisting of a list of objects in the environment and an instruction. The planner prompt conditions the LLM to break down the task into steps.
    \item At the second level, the LLM is prompted to perform two functions: one to calculate the action and another to check whether the current step of the plan is still valid. These functions involve further prompting of the LLM, with the instructions passed as function parameters.
\end{itemize}

The action prompt directs the LLM to calculate the required action based on the current state of the environment. The check prompt instructs the LLM to verify whether the instruction remains applicable in the current step of the generated plan. If the instruction no longer holds, an integer variable, available to the CodePolicy, is incremented. This variable, specific to the current episode, enables the CodePolicy to maintain a history of already-reached steps in the constructed plan. Both the variable and the state of the environment are accessible through an API. The prompts utilize few-shot prompting, with pairs of example instructions and answers. With CoT reasoning through hierarchical prompting, this approach reduces the reliance on the size of the LLM adding to its cost efficiency. Using code allows for the incorporation of third-party functions, broadening the complexity with which the LLM can reason about actions given the state of the environment. This approach enables the LLM to focus on reasoning about the task while outsourcing complex computations to external functions. Additionally, the process of generating a CodePolicy has minimal requirements, needing only a task description. This allows for easy adaptation to other tasks, similar to how human teachers can typically teach with just a brief description of a new task.

\paragraph{Providing Feedback} 
Evaluative feedback is given in LLM-iTeach to indicate that the agent's current action is beneficial for the task accomplishment and is provided as descriptive guidance. Corrective feedback is an action from the teacher to directly control the agent. To employ the LLM as a teacher, evaluative and corrective feedback is derived from the LLM's choice of action given the state of the environment. The action is determined by the CodePolicy and compared with the action predicted by the agent. Let the agent's action be \(a_{agent}\) and the CodePolicy's action be \(a_{llm}\). The actions are considered similar if the angle between them falls below a threshold, \(\beta\), as defined by the following equation:

\begin{equation}
\arccos \left( \frac{a_{agent} \cdot a_{llm}}{|a_{agent}|  \cdot |a_{llm}|} \right) \cdot \frac{180}{\pi} < \beta, \quad \beta \in [0^\circ, 180^\circ]
\label{direction}
\end{equation}

If Eq. \ref{direction} holds and the actions are similar, positive evaluative feedback is given. On the contrary, the LLM's choice of action is given as corrective feedback. In this way, it allows feedback to be provided in two forms, depending on the specific state and action.

\subsection{IIL Framework}

Oriented to that LLM-iTeach uses an IIL Framework that builds on the feedback generated by the LLM Teacher. Fig. \ref{fig:llm-teach} illustrates the interaction between the LLM Teacher and agent. At each time step, the teacher provides either corrective feedback $f_c$, which is executed instead of the agent's action, or evaluative feedback $f_e$, offering descriptive guidance on the agent's proposed action:

\begin{equation}\label{eq:Teacher}
    Teacher := H:\mathcal{S} \times \mathcal{A} \rightarrow \mathcal{F}, \quad \mathcal{F}:=\{f_e, f_c\}
\end{equation}

Evaluative feedback simply signals approval (\emph{good}), indicating the agent's action aligns with the teacher’s reasoning. In contrast, corrective feedback replaces the agent’s action, such that \(f_c \in \mathcal{A}\) is executed on the robot instead of the agent’s action at that time step. Actions executed on the robot are represented in a 4-dimensional space, consisting of a translational component and a binary gripper state. Negative evaluative feedback is not necessary, as corrective feedback is used in such cases. Since querying the LLM incurs a constant cost regardless of feedback type, this approach maximizes learning opportunities by providing corrective feedback in scenarios where disapproval would otherwise be indicated. This strategy enhances the amount of usable data, contrasting with methods that discard negative feedback in state-action pairs during training \cite{CEILing, DAGGER}.

The feedback is persisted for the agent to train its policy in parallel. From the policy, the agent determines its action for the next time step. For the agent, LLM-iTeach employs a stochastic policy based on the CEILing approach \cite{CEILing}, parameterized by a Gaussian distribution with the state-independent parameter $\sigma$:

\begin{equation}\label{eq:AgentCEILing}
    Agent := \mathit{\pi_\theta(a|s) \sim N(f_\theta(s,\theta);\sigma^2)}
\end{equation}

\begin{figure}
    \centering
    \includegraphics[width=\linewidth]{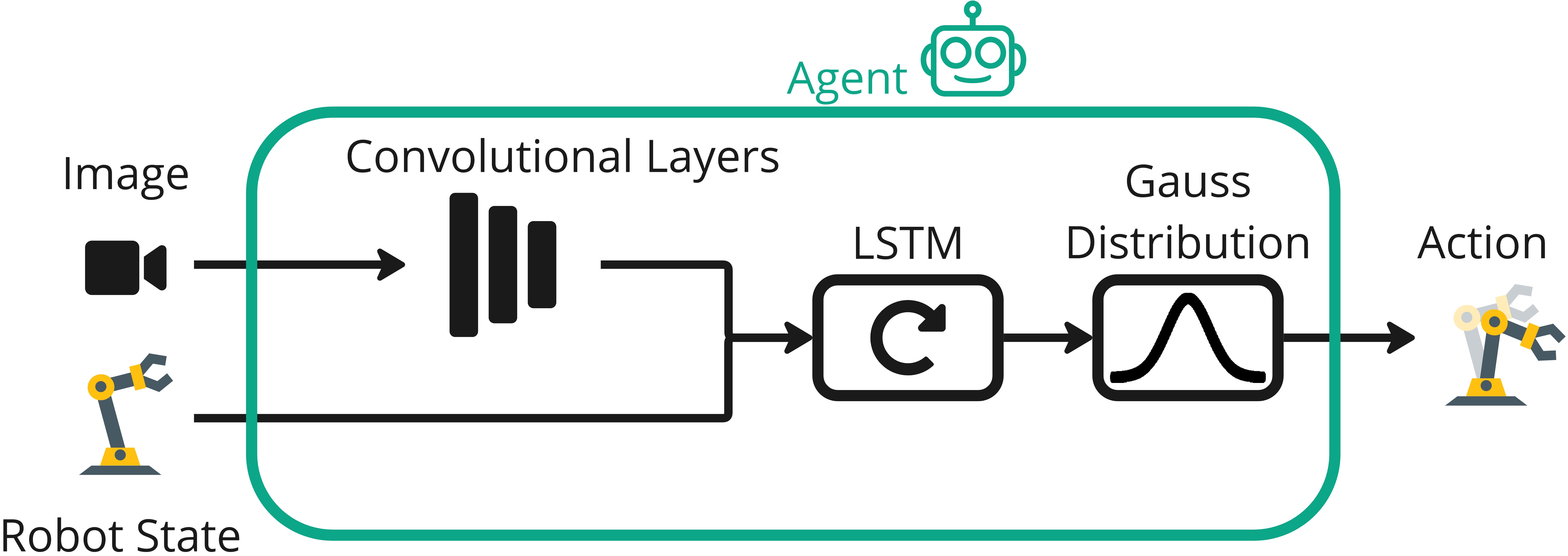}
    \caption{The model architecture for the learning agent in LLM-iTeach.}
    \label{fig:Agent}
\end{figure}

This formulation accounts for potential imperfections in the teacher's feedback, which may arise if corrective feedback dominates, yet the task cannot be completed. The agent’s policy, which incorporates evaluative feedback and a Gaussian distribution, induces exploration by introducing stochasticity into its decision-making process. The architecture of the agent's policy, shown in Fig. \ref{fig:Agent}, includes a camera attached to the robot arm to gather state information. 
 
\begin{figure*}[!ht]
    \centering
    \includegraphics[width=0.9\linewidth]{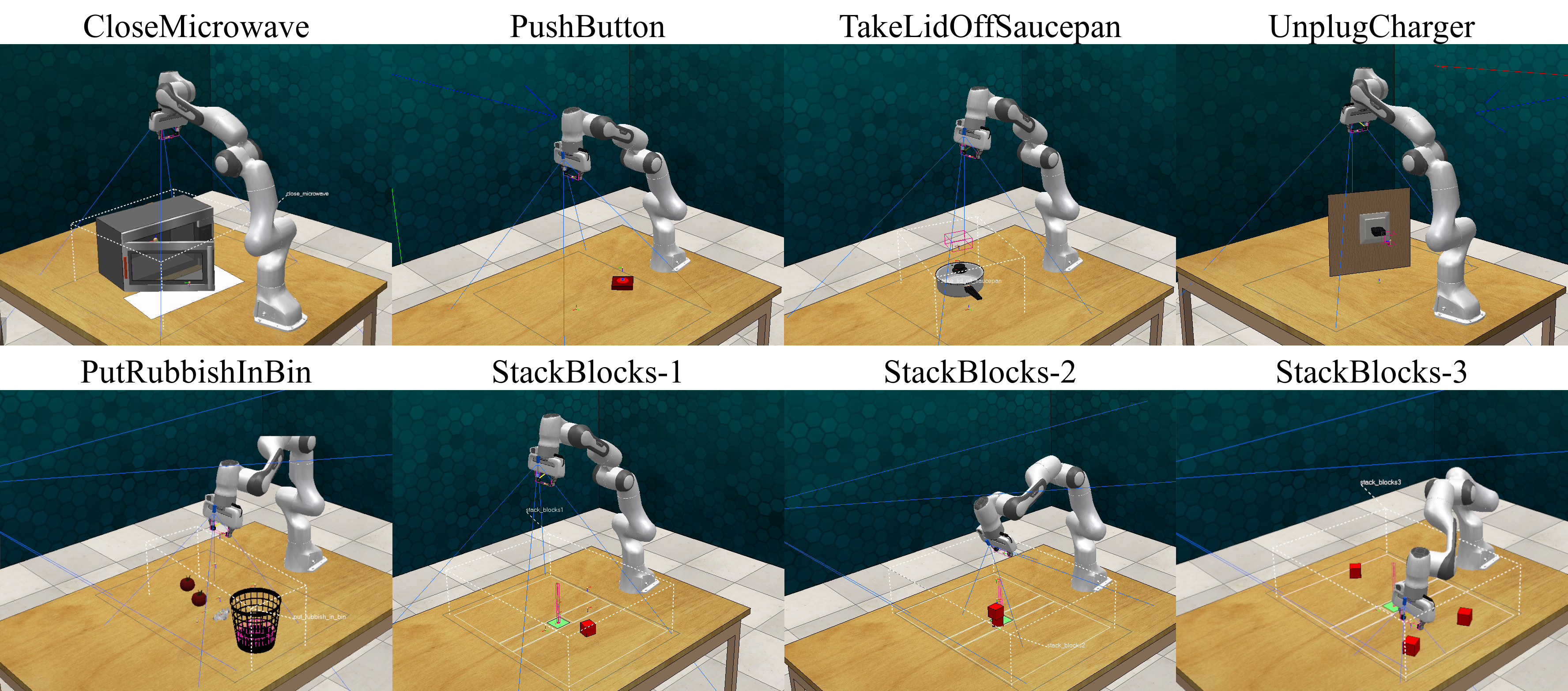}
    \caption{The tasks on the top row are shared with CEILing. The tasks on the bottom row are additional tasks. The screenshots are taken from the Graphical User Interface of RLBench \cite{RLBench}.}
    \label{fig:tasks} 
\end{figure*}


The agent learns its policy through an approximation of the objective function, optimized via the negative log-likelihood of the policy under the Gaussian distribution. The loss function is defined over the state-action pairs generated by interactions with the environment and the teacher’s feedback:

\begin{equation}\label{eq:lossCEILing}
    L(\pi,a) = -q\log(\pi_\theta(a|s))
\end{equation}

The scalar \(q\) weights each state-action pair according to the feedback \(f\) provided at time step \(t\), inspired by the IWR method \cite{IWR}. Specifically, \(q\) is defined as:

\begin{equation}\label{eq:q}
q_t =
\begin{cases} 
1 & \text{when } f_e \text{ is given} \\
N/N_c & \text{when } f_c \text{ is given}
\end{cases}
\end{equation}
where \(N\) represents the total number of time steps in an episode, and \(N_c\) denotes the number of corrective feedback actions given during that episode. When $f_e$ is given, the action $a$ is taken from the agent's policy, while $f_c$ is given in terms of another action $a$ following the LLM's policy. This weighting system emphasizes corrected state-action pairs, ensuring that their importance is not diminished even if a majority of the feedback is corrective. In scenarios where corrective feedback dominates, the weighting of corrective actions is gradually adjusted to align with actions that received positive evaluative feedback, thereby maintaining balanced learning. 
Unlike human teachers who are sufficiently knowledgeable to finish the task, the LLM's policy lacks consideration of Inverse Kinematics or the actual dynamic environment, like the sizes of objects or spatial relationships between them. Therefore, episodes that exceed a predefined number of time steps \(N_t\) are aborted, and their state-action pairs are excluded from the learning dataset.

\begin{figure*}[!ht]
    \centering
    \includegraphics[width=0.9\linewidth]{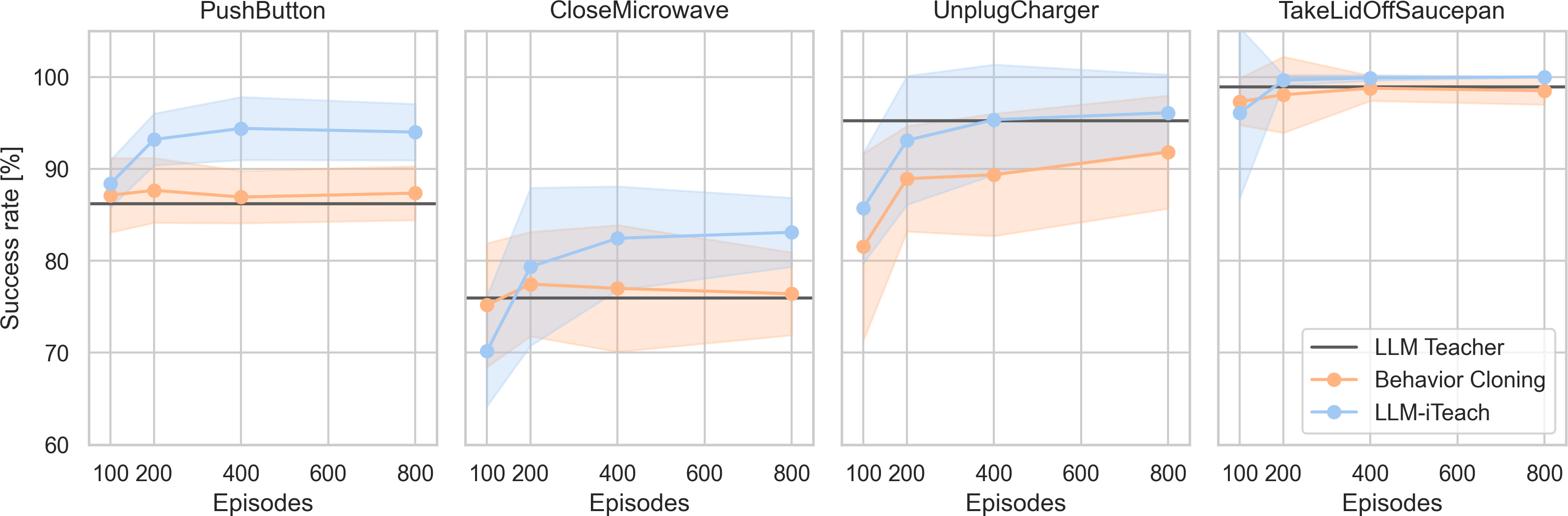}   
    \caption{The average success rate of BC and LLM-iTeach for the number of episodes used in the training phase. Additionally, the performance of the LLM Teacher is given.}
    \label{fig:LLMTeachBCAll}
\end{figure*}

\makeatletter
\begin{figure*}[htbp]
    \renewcommand{\@captype}{table}
\begin{center}
    \centerline{\includegraphics[width=\linewidth]{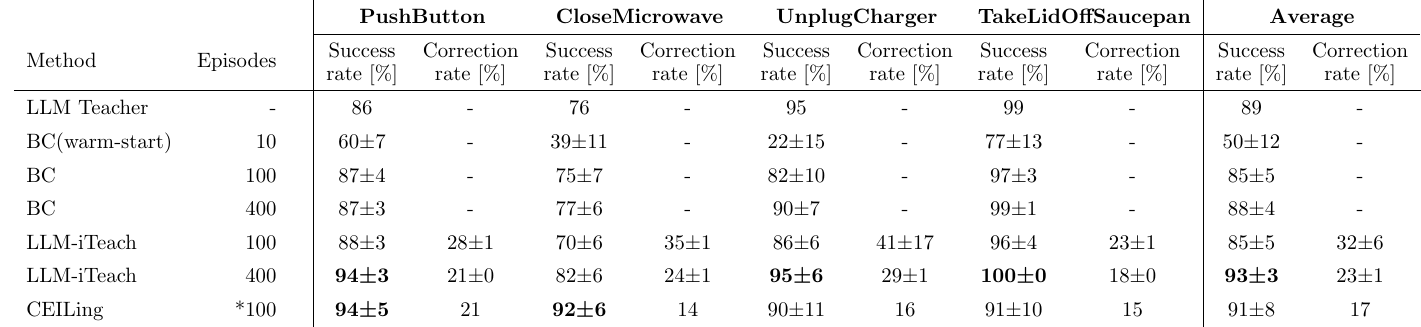}}
\end{center}
\vskip -0.2in
\caption{Results for LLM-iTeach, BC, CEILing, and Warm-start with the number of episodes used in the training and the standard deviation. As a reference, the success rate of directly letting the LLM control is given as LLM Teacher (also depicted in Fig. \ref{fig:LLMTeachBCAll}). The * indicates that the episodes were collected with feedback from a human teacher. The correction rate is the average rate of corrective feedback to all feedback. The results from CEILing are taken from its publication \cite{CEILing}.}
\label{tab:results}
\end{figure*}
\makeatother




\section{Experiments}\label{sec:experiments}
To evaluate to what extent an LLM can interactively teach an IL agent in robotic manipulation by providing corrective and evaluative feedback, we apply LLM-iTeach on various RLBench \cite{RLBench} tasks in the CoppeliaSim\footnote{https://www.coppeliarobotics.com/} simulator, in which a simulated Franka Emika Panda robot with 6 degrees of freedom and a gripper as an end-effector set on a worktable. To be specific, LLM-iTeach's performance is compared against the baselines of self-imitation with Behavior Cloning (which belongs to IL) and CEILing (which belongs to IIL) to analyze the LLM's relative effectiveness in teaching. In addition, the ablation study regarding the integration of two types of feedback and the hyperparameter study for $\beta$ are reported accordingly.

\subsection{Setup}
To attain comparability, we first keep the same experimental setup with CEILing \cite{CEILing} by employing BC and LLM-iTeach in four RLBench tasks for robot manipulation (displayed in the top row of Fig. \ref{fig:tasks}). Then, to evaluate the extension ability of LLM-iTeach, we further employ LLM-iTeach in four additional tasks (displayed in the bottom row of Fig. \ref{fig:tasks}). Among all the tasks, each method involves a training phase, and an evaluation phase with 100 episodes to measure the success rate. Each experiment is repeated at least 20 times per task for statistical robustness. The experiments of CEILing, we want to compare LLM-iTeach to, are limited to 100 episodes each, considering the human effort in providing feedback. With the advantage of cost-effective scalability, we conduct our experiments in addition to the limit of 100 with 200, 400 and 800 episodes. These experiments are relatively inexpensive to produce compared to methods that rely on a human teacher and offer a way to analyze the impact on performance, trough scaling. For comparability, we also repeat the scaled experiments with the BC baseline where we train on these different numbers of demonstrations. The LLM is realized by Llama3-70b\cite{dubey2024llama} and leverages an API that provides object positions and robot states for grounding. Inverse kinematics from RLBench is applied to map actions to the robot's joint movements. The LLM Teacher is also utilized without the agent and in direct control of the robot to collect demonstrations for warm-starting the agent’s policy. This is done for all methods, where the agent's policy is initialized with 10 demonstrations. These demonstrations, referred to as warm start, are additionally inspected as the only input to the agent, to see the improvements of the methods above its initial state. The LLM is prompted with a temperature setting of 0.0, and interactions between the agent and teacher, which involve the creation of feedback, occur at a frequency of 20 Hz. The agent's configuration has the parameter $\sigma$ in Eq. \ref{eq:Agent} set to 1 mm for translations.




\subsection{Overall Results}\label{results}



The performance is compared to the baselines of CEILing and BC. It shows that on average LLM-iTeach performs the best with similar success rates to CEILing, while BC has on average a 5 \% lower success rate than LLM-iTeach. Although LLM-iTeach is only surpassed by CEILing in the individual task CloseMicrowave, the overall results indicate that LLM-iTeach is a viable method for training an agent to complete a robotic task. It exceeds the average performance of BC and matches that of CEILing. This demonstrates that LLM-iTeach is well-suited to address the given task complexity compared to other methods, without the constraints associated with human involvement, which are related to its higher cost and time consumption. 


In Tab. \ref{tab:results}, the task-specific results are given with LLM-iTeach and BC trained on 400 and 100 demonstrations to compare to CEILing which only uses 100. With 100 demonstrations LLM-iTeach and BC performed worse on average. As the number of episodes with the LLM Teacher does not face a cost limit relative to using a human teacher, setting the number of training episodes for LLM-iTeach and BC to 400 is considered feasible. The number of 400 is determined by the experimental results as the method does not improve significantly in performance from that number, as seen in Fig. \ref{fig:LLMTeachBCAll}. The same applies to the amount of demonstrations used for BC. Also, we can observe a drop in the correction rate with a higher number of episodes, being another indicator for the learning of the agent, as its predictions get relatively more evaluative feedback and require less corrective intervention by the teacher. LLM-iTeach with 400 demonstrations gives both corrective and evaluative feedback with a similar ratio to CEILing, contributing to the isolation of the teachers' performance. Furthermore, the table displays the warm-start policy as a reference for LLM-iTeach and CEILing, since both methods utilize the warm-start at the start of the task.


It can be observed that BC reaches its performance ceiling at 200 demonstrations, as increasing the number to 400 or 800 does not lead to any significant improvement in success rate. A comparison between BC and LLM-iTeach reveals that BC's performance plateau is comparable to that of the LLM Teacher across most tasks, while LLM-iTeach not only surpasses BC but also outperforms the LLM Teacher in the majority of tasks. This provides empirical support for the hypothesis that incorporating exploration through evaluative feedback from the agent's policy is beneficial. The performance gains achieved by LLM-iTeach also justify its additional complexity compared to BC. During training and evaluation, we noticed that when an episode failed, it was often due to the inverse kinematics algorithm being unable to execute the desired actions on the robot. Since both the LLM and, by extension, the CodePolicy rely solely on the API and lack external contextual knowledge, LLM-iTeach does not consider such errors under the physical constraints of the 3D world. This limitation contributes to the reduced success rate of our method, especially in the CloseMicrowave task, in which the LLM intends to interact with the microwave door handle, despite the fact that pushing any part of the door surface would be sufficient to close it. This behavior underscores a gap in the LLM's understanding of physical constraints, in contrast to a human teacher who intuitively grasps the physical laws of the real world.

\subsection{Ablation Studies} \label{sec:ablation_studies}

\begin{figure}{}
    \centering
    \includegraphics[width=0.95\linewidth]{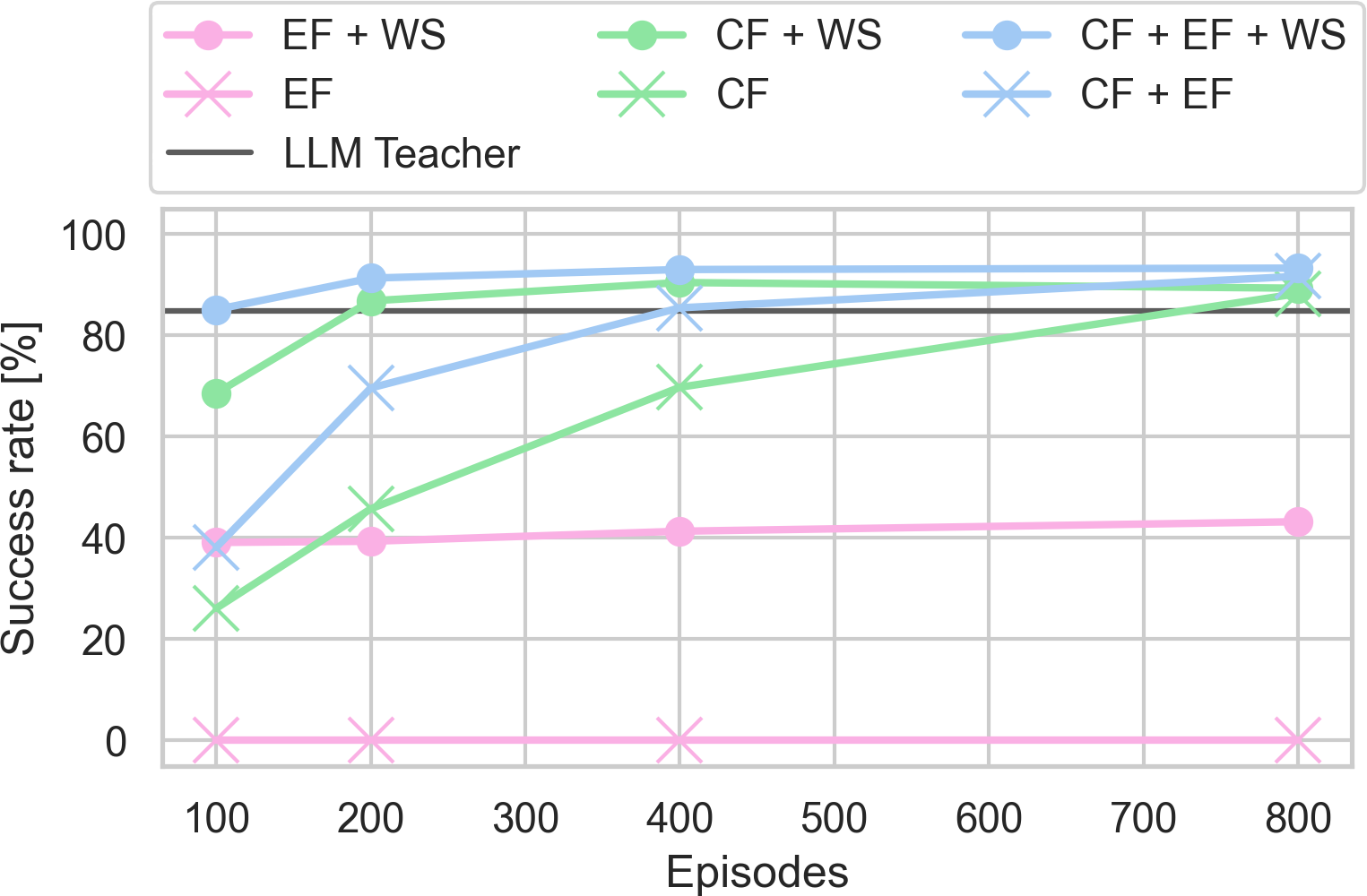}
    \caption{Average success rate of LLM-iTeach over a varying number of episodes in the training phase averaged over all 4 tasks. CF: corrective feedback applied; EF: evaluative feedback applied; WS: warm-start used.}
    \label{fig:ablation}
\end{figure}

The results of the ablation studies are depicted in Fig. \ref{fig:ablation} for the different cases with only evaluative feedback, only corrective feedback, and for the combination respectively. In all three instances, the performance is given with/without a warm-start policy. The success rates are averaged over all tasks and are shown to the number of episodes used during the training phase. For LLM-iTeach with only evaluative feedback, the method was not able to achieve any successful task completions with the agent, if no warm-start was applied in the training phase. In contrast, with the warm-start, the method could train an agent, that could complete the task, but its performance did not increase significantly over a longer training phase. Its performance is similar to BC's with the same number of demonstrations as the warm-start (see Tab. \ref{tab:results}), suggesting that this ablation does not teach the agent any more than it learned from the warm-start. In the ablations involving corrective feedback the performance without warm-start is lower than with warm-start and gradually reaches the same level with a rising number of episodes during the training phase. The same can be said when both feedback types are applied. The method with the warm-start seems to have a performance limit at around 400 episodes in the training phase, whereas the method without the warm-start has the same limit at double the length. This shows that initializing the agent's policy with the warm-start shortens the required length of the training phase significantly as the warm-start only requires 10 demonstrations. Overall, the combination of both feedback types and warm-start, the policy achieves the highest performance and corresponds to LLM-iTeach.

Comparing only corrective feedback to both feedback types shows that the combination results in a higher success rate for the agent. It is even higher than that of the teacher, while only corrective feedback has a success rate similar to that of the teacher. This further supports the assumption that the evaluative feedback to the agent induces exploration with its policy.

\subsection{Hyperparameter Study} \label{sec:hyperparameter_study}
\begin{figure}
\centering
    \includegraphics[width=0.95\linewidth]{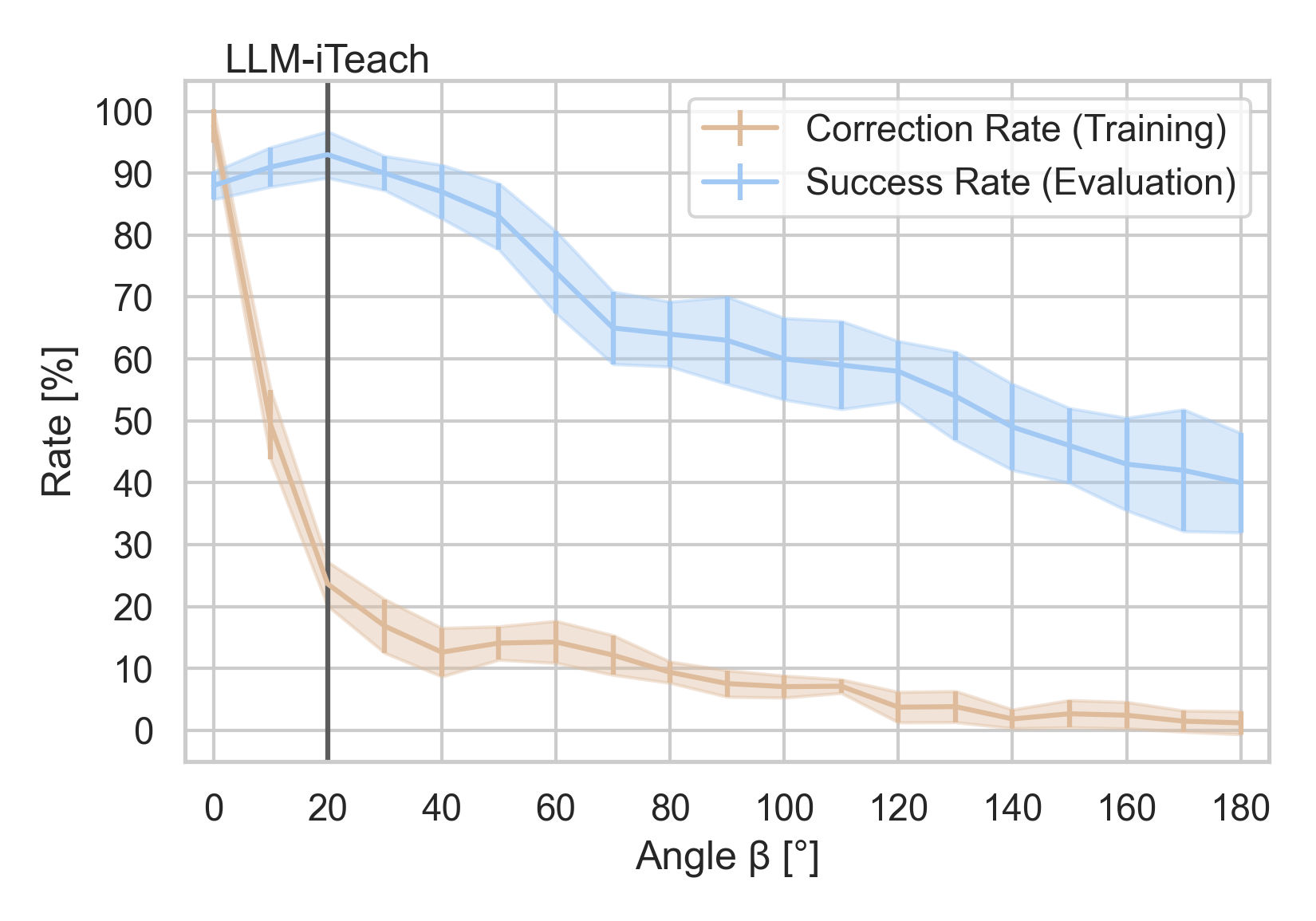}
    \caption{Hyperparameter study on different threshold angles. The success rate in the evaluation phase and the rate of evaluative to corrective feedback given in the training phase against different values of the hyperparameter $\beta$. The value determined for LLM-iTeach is marked with a vertical line.}
    \label{fig:hyper}

\end{figure}

\makeatletter
\begin{figure*}[]
    \renewcommand{\@captype}{table}
\begin{center}
    \centerline{\includegraphics[width=0.95\linewidth]{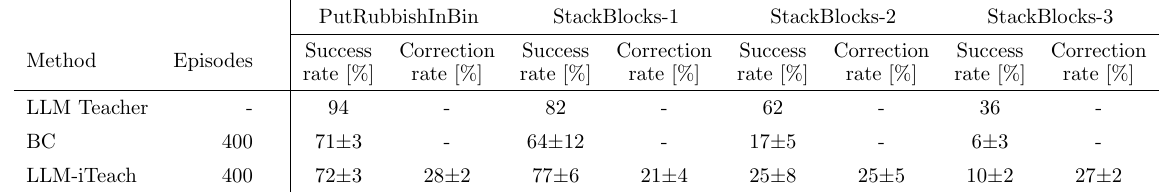}}
\end{center}
\vskip -0.2in
\caption{Results from the experiments with additional tasks for LLM-iTeach. Baseline experiments with BC are given, as well as the LLM's performance in directly executing all actions.}
\label{tab:results_additional_tasks}
\end{figure*}
\makeatother

The hyperparameter study focuses on the threshold parameter $\beta$ for determining the similarity of the teacher's and agent's actions. The results presented in Fig. \ref{fig:hyper} show the agent's performance and rate of corrections in the teacher's feedback trained on 400 episodes averaged across all four tasks, with $\beta$ varying between $0^\circ$ and $180^\circ$. When $\beta = 0^\circ$, the success rate is that of LLM-iTeach with only corrective feedback, while the result in a situation when $\beta = 180^\circ$, the result mirrors the performance of LLM-iTeach with exclusively evaluative feedback, as observed in section \ref{sec:ablation_studies}. Although the evaluative data is subject to noise due to the limited sample size, we found that the highest success rate occurs when $\beta$ is set to 20 degrees. At this value, the agent's performance is, on average, comparable to that of the CEILing framework (as shown in Tab. \ref{tab:results}). We also observe that for the best performance, we reach a balance of corrective and evaluative feedback, leveraging the information gained from both types. As $\beta$ increases, the success rate declines, reaching its lowest point at 180°, where the correction rate approaches 0\%. To summarize the findings in Fig. \ref{fig:hyper}, a balance between corrective and evaluative feedback leads to the best performance in the shortest time. As the LLM Teachers cannot fully observe the task environment in our setup, this balance accounts for possible resulting suboptimal corrective feedback.




\subsection{Results for Additional Tasks}

We verify the generalization of our method by testing it on an additional four tasks. The results are depicted in Tab. \ref{tab:results_additional_tasks}. It spans the performance of BC as the baseline, the LLM Teacher, and LLM-iTeach. LLM-iTeach exceeds the success rate of BC in every task. The results reflect the capability of LLM-iTeach to also learn long-horizon tasks. However the 20\% better performance of using the LLM Teacher to directly act on the robot indicates potential in the learning architecture that is not fully leveraged. Using the LLM Teacher directly to utilize this better performance would conflict with the potential of applying it to real-world scenarios where privileged information (e.g., ground truth states) may not be accessible during training. Nevertheless, the performance difference shows that using the LLM with prompting for the Codepolicy works well even if multiple steps are required. Similar to the other four tasks, we observed that the CodePolicy mastered a wide range of sequential decision-making to complete the tasks. Yet the error of the inverse kinematics algorithm, together with the lack of knowledge outside the provided information by the API, resulted in frequent failure to complete the task. The expansion to the additional tasks required the LLM Teacher to prompt for CodePolicy from only a simple descriptive sentence of the tasks, so it came at a small cost, portraying the easy extendability of LLM-iTeach.  



\section{Conclusion \& Discussion}\label{conclusion}

This work introduces LLM-iTeach, an IIL framework that leverages LLMs instead of human teachers to train agents in solving robotic manipulation tasks. Firstly, LLM-iTeach utilizes a hierarchical prompting architecture to generate a CodePolicy for the task, then with the designed similarity-checking mechanism, enables the LLM to teach the agent with corrective and evaluative feedback. 

LLM-iTeach trains the agents over a variety of sequential decision-making tasks with a higher success rate compared to the baseline of BC and performance similar to that of the state-of-the-art IIL method CEILing. Without the extensive need for human resources during training, this establishes LLM-iTeach as a viable and cost-reducing alternative. Notably, our results suggests that LLMs can teach agents as effectively as humans. In the evaluation, the teaching capabilities of both the human and LLM were validated by applying the same experimental setup in LLM-iTeach that was used in CEILing, allowing us to isolate the respective contributions of both teachers to the overall performance of the methods. While the training time may be longer with LLMs, their lower cost and constant availability make LLM-iTeach a practical choice for applications where these factors are critical. 

Using evaluative feedback in combination with corrective feedback in LLM-iTeach allows the agent to better adjust its policy based on cumulative performance rather than just immediate corrections, making it particularly useful in complex, sequential tasks. While evaluative feedback is often associated with reinforcement learning, in this case, it complements corrective feedback by providing a more holistic view of the task outcome. Corrective feedback alone can lead to suboptimal behavior as it focuses only on immediate errors, whereas the inclusion of evaluative feedback enhances the agent's ability to refine its overall strategy. Introducing exploration into the agent's policy when receiving evaluative feedback allowed the agent to outperform the teacher in some cases, accounting for some of the limitations of the LLM Teacher.

The limitations of LLM-iTeach are 1) The LLM Teacher's restricted observation space makes the LLM policy lack fine-grained consideration of the robotic task environment; 2) The limited action space of LLM-iTeach, while achieving considerable performance to state-of-the-art methods, leaves potential for more complex policies 3) The reliance on ground truth data, which is required for the presented setup of LLM-iTeach to generate accurate feedback, limits the approach's applicability to real-world scenarios. Future work could address these limitations in several ways. Expanding the LLM Teacher's observation space, for instance, by incorporating vision-language models, could provide more versatile feedback and potentially generate better demonstrations, as seen in related research. Moreover, relying less on ground truth data, advancing LLM-iTeach to a 7-d action space, and moving towards real-world sensory inputs will enhance the performance and practicality of this approach. 
\section{Acknowledgment}
The authors gratefully acknowledge support from the China Scholarship Council (CSC) and the Horizon Europe project TERAIS (Grant No. 101079338). We also extend our sincere thanks to Josua Spisak for his valuable feedback on the manuscript.


\bibliographystyle{ieeetr} 
\bibliography{reference}

\end{document}